\documentclass{article}
\usepackage{arxiv}
\usepackage[utf8]{inputenc} % allow utf-8 input
\usepackage[T1]{fontenc}    % use 8-bit T1 fonts
\usepackage{hyperref}       % hyperlinks
\usepackage{url}            % simple URL typesetting
\usepackage{booktabs}       % professional-quality tables
\usepackage{amsfonts}       % blackboard math symbols
\usepackage{nicefrac}       % compact symbols for 1/2, etc.
\usepackage{microtype}      % microtypography
\usepackage{lipsum}
\usepackage{bbold}
\usepackage{bold-extra}
\usepackage{lmodern}
\usepackage{amsmath}
\usepackage{amsthm}
\usepackage{amsfonts}
\usepackage{gensymb}
\usepackage{mathtools}
\usepackage{subfig}
\usepackage{graphicx}
\usepackage{float}
\usepackage{verbatim}
\usepackage{caption} 
\newcommand{\fTwo}{f(X|\theta)}
\newcommand{\f}{f(X|\theta_{i})}
\newcommand{\g}{g(X|\phi_{i})}
\newcommand{\h}{h(X|\xi_{i,j})}
\DeclareMathOperator*{\argmax}{argmax}

\newbox{\myorcidaffilbox}
\sbox{\myorcidaffilbox}{\large\includegraphics[scale=0.005]{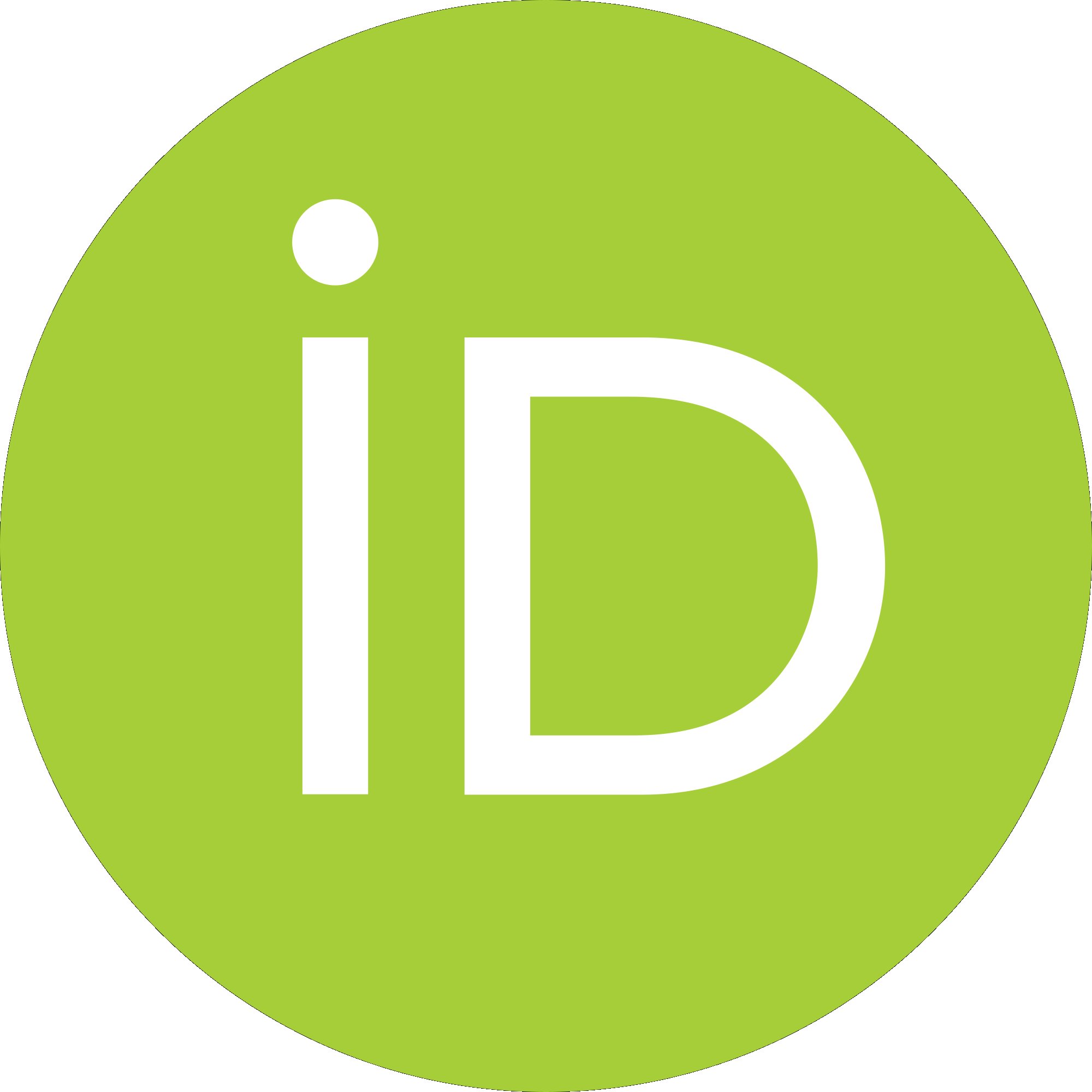}}
\newcommand{\orcidaffil}[1]{%
  \href{https://orcid.org/#1}{\usebox{\myorcidaffilbox}}}

\title{Topological Navigation Graph Framework\thanks{
	This is a preprint of an article published in Autonomous Robots. The final authenticated version is available online at: \href{https://doi.org/10.1007/s10514-021-09980-x}{https://doi.org/10.1007/s10514-021-09980-x}.
	}}

\author{
	Povilas Daniu\v{s}is\thanks{Corresponding author: Email: povilasd@neurotechnology.com} \hspace{1mm}\orcidaffil{0000-0001-5977-827X}, \hspace{0.5mm} %\\
	Shubham Juneja \hspace{0.5mm}\orcidaffil{0000-0002-7906-5688},\hspace{0.5mm}%\\
	Lukas Valatka\thanks{Worked in Robotics department at Neurotechnology during the research.} \hspace{1mm}\orcidaffil{0000-0002-4734-6442},\hspace{0.5mm}%\\
	Linas Petkevi\v{c}ius \hspace{0.5mm}\orcidaffil{0000-0003-2416-0431}\\
	Neurotechnology \\
}

\begin{document}
\maketitle
\begin{abstract}

  We focus on the utilisation of reactive trajectory imitation controllers for goal-directed mobile robot navigation. We propose a topological navigation graph (TNG) - an imitation-learning-based framework for navigating through environments with intersecting trajectories. The TNG framework represents the environment as a directed graph composed of deep neural networks. Each vertex of the graph corresponds to a trajectory and is represented by a trajectory identification classifier and a trajectory imitation controller. For trajectory following, we propose the novel use of neural object detection architectures. The edges of TNG correspond to intersections between trajectories and are all represented by a classifier. We provide empirical evaluation of the proposed navigation framework and its components in simulated and real-world environments, demonstrating that TNG allows us to utilise non-goal-directed, imitation-learning methods for goal-directed autonomous navigation.
  
\end{abstract}
\section{Introduction}

Combining perception and action into a single algorithm of fundamental and practical importance, mobile robot navigation is an attractive area for artificial intelligence research. 
Despite significant efforts, both from academia and industry, this area still poses challenges that need to be resolved in order to create autonomous systems capable of operating efficiently in real-world environments. 
For many years, the mainstream direction of robot navigation research has been largely focused on methods relying on accurate sensors and direct (e.g. probabilistic-geometric) models~\cite{thrun2005probabilistic,ORBSLAM,LSDSLAM,Cadena16tro-SLAMfuture}.
Regardless of certain advantages (e.g. the interpretability of geometric maps), this approach poses some serious limitations, such as a lack of flexibility, drastically increasing complexity when large numbers of situations have to be identified and handled by the model.

On the other side, when humans learn various complex sensorimotor behaviours, including the skill to navigate, adaptivity and imitation play crucial roles~\cite{gershman2015computational}.
Once children learn several basic behaviours, they soon figure out how to connect them in order to reach different, increasingly complex goals~\cite{calero2019language}. 
Biological neural networks are excellent learners from examples. For more than three decades, artificial neural networks have been utilised to build autonomous mobility systems or their components ~\cite{Pomerleau1988ALVINNAA,GaussierPerAc,LeCun:04,BojarskiTDFFGJM16,8100252,banino2018vector,SavinovDosovitskiyKoltun2018_SPTM}.
The recent progress of neural network methods, fuelled by deep learning techniques, has been successfully adopted by mobile robot autonomy researchers, resulting in several important achievements in this field. 
For example,~\cite{BojarskiTDFFGJM16} applies behaviour cloning and modern convolutional neural networks (CNNs), achieving vehicle autonomy on real roads,~\cite{bewley2019sim2real} describes a transfer learning architecture for learning to drive a real vehicle solely in a simulator,~\cite{CodevillaMLKD18} contributes a neural architecture for conditional imitation learning,~\cite{pathakICLR18zeroshot} introduces a method to learn navigation policies from self-supervision,~\cite{Bansal2018ChauffeurNetLT} proposes a recurrent neural network-based solution with a demonstration using real cars. 
These works, in various aspects, also rely on imitation learning, which highlights a tendency that, by benefiting from the current renaissance of connectionist systems, imitation learning is becoming an increasingly popular robot autonomy research direction~\cite{Osaetal18}.

% (Povilas)
%  Highlighted problem we try to solve with TNG.

% Problem formulation and state of the art limits.
One of the most attractive properties of the imitation learning approach is the possibility of learning complex sensorimotor behaviours from demonstration data instead of direct programming.
However, many imitation learning algorithms (e.g.~\cite{BojarskiTDFFGJM16,bewley2019sim2real,NIPS2016_6391}) are not goal directed by design. 
This does not allow practitioners to use these algorithms in applications, which require goal-directed behaviour. 
In this article, we solve this problem by suggesting a topological navigation graph (TNG) framework, which allows one to create goal-directed navigation systems from non-goal-directed, imitation-learning components.

The main contributions of this study include the TNG framework, the application of neural object detection architectures to obtain more robust trajectory imitation, and the experimental evaluation of the suggested algorithms both in a simulator and the real world by using Neurotechnology's SentiBotics mobile robot development kit~\cite{sentibotics_navigation_sdk}.

\begin{figure*}[t]
  \begin{center}
  \includegraphics[scale=0.280]{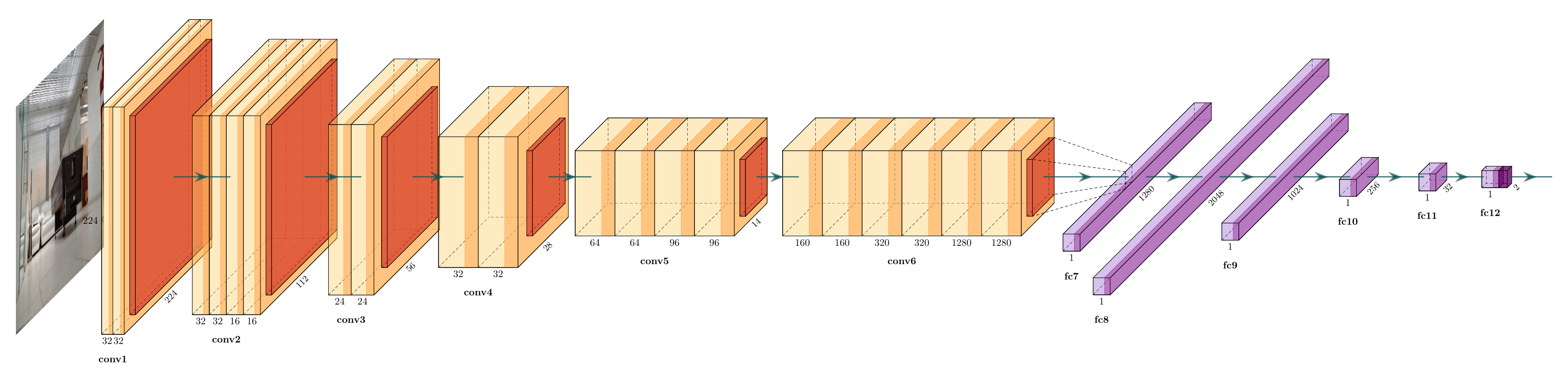}
  \caption{The architecture of regression controller~\eqref{eq:controller_representation1}, represented by the composition of pre-trained CNN (Mobilenet V2) and trainable MLP.}\label{fig:mobilenet}
  \label{fig:mobilenet}
  \end{center}
\end{figure*}

The rest of paper is organised as follows.
Section~\ref{sec:trajectory_controllers} describes the core component of TNG - trajectory imitation controllers. Therein, we describe regression controller and propose a new, object-detection-based trajectory imitation controller. Section~\ref{sec:topological_planning} is devoted to TNG and its implementation, and Section~\ref{sec:experiments} provides an empirical evaluation of the proposed algorithms.
Finally, Section~\ref{sec:conclusion} concludes the article.

\section{Trajectory Imitation Controllers}
\label{sec:trajectory_controllers}

In this section, we focus on learning a reactive skill of visuomotor trajectory following independently from the overall navigation method. 
As becomes apparent in Section~\ref{sec:topological_planning}, this independence adds modularity facilitating in performing complex behaviours as a whole, where every reactive task is a building block.

The trajectory following skill is represented by the trajectory imitation controller $y_{t} = f(X_{t}|\theta)$, where $X_{t} \in \mathbb{R}^{H\times W \times 3}$, $y_{t}\in \mathbb{R}^2$, and $\theta  \in \mathbb{R}^d$, respectively denote camera image inputs (3 channel RGB image, with height H and width W), motor command outputs, and learnable parameters of the parametric function $f$ (e.g. a deep neural network). 
In order to obtain the trajectory imitation controller, the robot is first driven by an expert while the corresponding camera image and motor command data pairs $\big(X_{t}, y_{t} \big)_{t=1}^{n}$ are collected.
The data pairs are utilised to optimize the learnable parameters $\theta$ of the neural network. 
Once the parameters have been estimated, the robot is expected to be able to follow the learned trajectory.

The problem of learning imitation controllers has a long history~\cite{Pomerleau1988ALVINNAA} and is currently being actively researched~\cite{BojarskiTDFFGJM16,7995721,CodevillaMLKD18,pathakICLR18zeroshot,Wang2018DeepOC,bewley2019sim2real}.
Although various solutions have been proposed, in general, it remains unsolved both in the fundamental and technical senses.
Important challenges associated with the trajectory following problem are related to natural environmental variations, such as lighting conditions, changes in certain environmental elements (e.g. objects like furniture), as well as the appearance of obstacles, pedestrians, and other dynamics. In most cases, it is not technically feasible to directly encode all the environmental variation into training sets of acceptable size. Therefore, we seek models that can generalise and distil the desired behaviour from realistic data sets. 

We compare two models for this task. Our first model is derived from~\cite{BojarskiTDFFGJM16} which regresses motor commands with the use of a CNN, followed by our second model which is based on an object detection neural network.

\subsection{Regression Controller}
\label{sec:trajectory_controller_cnn_mlp}

Trained according to behaviour cloning~\cite{Osaetal18}, the regression controller is a neural network model:

\begin{equation}
 \fTwo = f_{\text{MLP}}( f_{\text{CNN}}(X|\hat{\xi})| \theta),
 \label{eq:controller_representation1}
\end{equation}

\noindent where $f_{\text{MLP}}$ is a multilayer perceptron (MLP)~\cite{murphy2013machine}, and $f_{\text{CNN}}$ is fixed a CNN encoder (e.g. $\text{ResNet}$~\cite{ResNet} or $\text{MobilenetV2}$~\cite{MobilenetV2}), with fixed weights $\hat{\xi}$ pre-trained on $\text{ImageNet}$~\cite{deng2009imagenet} classification task, and $\theta \in \mathbb{R}^d$ represent trainable parameters.
Having a training set we obtain parameter estimate $\hat{\theta}$ by minimising the loss function of the ridge regression~\cite{friedman2001elements}:

\begin{equation}
L(\theta) = \sum_{i} ||y_{i} - f(X_{i}| \theta)||^{2} + \lambda ||\theta||^{2}, \:\:\:\:\:\: \text{where}\: \lambda > 0.
\label{eq:cost_function1}
\end{equation}

We optimise~\eqref{eq:cost_function1} with the Adam optimisation algorithm~\cite{Adam}.
As compared to similar work by~\cite{BojarskiTDFFGJM16}, where the entire parameter set is optimised, our approach helps to reduce the training duration without affecting the performance. %and rules out the need for large amounts of training data.
This may be due to the fact that pre-trained CNN encoders extract rather general features, which are efficient representations of points in trajectory.
Following the insights of~\cite{CodevillaMLKD18} we also use dataset aggregation (DAgger~\cite{DAgger}) and augmentation (input/output shifting, random lighting, and regional dropout), allowing us to improve the model's stability and robustness.

% maybe worth move to Section 4?
We empirically identified several limitations in the aforementioned approach. 
Firstly, controllers trained with this approach are highly sensitive to rotational errors; that is, slight deviations from the trajectory may cause cascading errors~\cite{Osaetal18} and eventually drive the robot out of the trajectory with no possibility of self-recovery. 
We used input/output shifting data augmentation to solve this problem, but it only helps to some extent. 
The second limitation is the model's sensitivity to changes in the environment, which occur due to illumination variations, moved furniture, pedestrians, and so forth. 
To tackle this problem, additional iterations of DAgger can be performed, but with further unavoidable changes in the environment, the problem reoccurs. 
Thirdly, the trajectory following performance during execution does not necessarily correlate with the value of offline evaluation over the test set using the loss function~\cite{Codevilla2018OnOE}, hence complicating the model evaluation.

\subsection{Object Detection-Based Controller}
\label{sec:od_trajectory_controllers}

In order to solve the aforementioned problems, we utilize the RetinaNet object detection architecture~\cite{Lin_2017_ICCV}.
Conceptually, our approach is similar to that of~\cite{Wang2018DeepOC}.
However, instead of detecting specific objects, which may be not present in the navigation scene, we train the object detector to detect the robot's direction of movement, represented with a single class bounding box on the input image.

\begin{figure}[H]
	\centering
	\subfloat[steering: LEFT]{\includegraphics[width=0.25\textwidth]{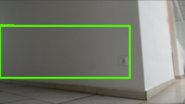}}
	\subfloat[steering: RIGHT]{\includegraphics[width=0.25\textwidth]{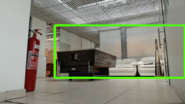}}  

	\subfloat[steering: NONE]{\includegraphics[width=0.25\textwidth]{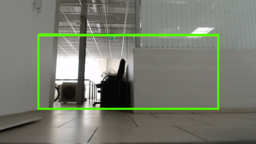}}

	\caption{Samples of a training set as per rotation axis.}
	\label{img:traj_detection_inputs}
\end{figure}

The training data consist of camera images paired with bounding box coordinates as per the rotation axis of the robot in the image.
The bounding-box are centered if the robot is being driven straight, and horizontally shifted in the direction of the rotation if the robot is rotated (see Figure~\ref{img:traj_detection_inputs}).

%We observed shifting to be necessary for successful trajectory following at points where the robot has to make sharp turns (e.g. when steering around a corner).
The model used for training is pre-trained on the COCO dataset~\cite{lin2014microsoft} over the object detection task.
We optimise all parameters using stochastic gradient descent with momentum optimisation algorithm.

\begin{figure}[H]
	\centering    \includegraphics[width=0.25\textwidth,height=\textheight,keepaspectratio]{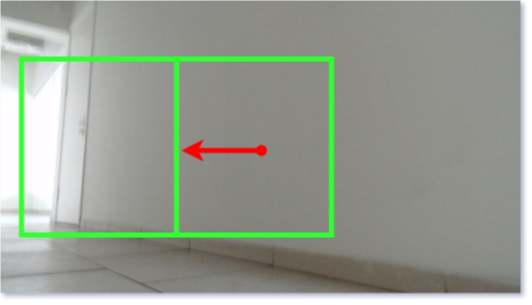}
	\caption{Predicted bounding box and evaluated command  (arrow) to turn.}
	\label{img:traj_detection_error}
\end{figure}

During controller execution, the output of the object detector (i.e. coordinates of the bounding box showing the direction of movement) is coupled with a PID controller~\cite{astrom2010feedback} to centre a detected bounding box, resulting in trajectory imitation.
The error input for the PID controller is defined as $e_{t}:=x_{t}-x_{c}$, where $x_{t}$ and $x_{c}$ are the detected bounding box's centre $x$ coordinate and the centre $x$ coordinate of the input image (see Figure~\ref{img:traj_detection_error}).

Comparing to the approach of Section~\ref{sec:trajectory_controller_cnn_mlp}, an object detection-based controller, by design, allows recovery from larger rotational errors, as learnt bounding boxes can be detected from various geometric transformations of training examples without any recovery-aiding data augmentations. 
We observed that this controller has a tendency to produce detections with low confidence for images far outside the distribution of the training sets, which adds the possibility of avoiding potentially incorrect decisions.

\section{Topological Navigation Graph (TNG)}
\label{sec:topological_planning}

In this section, we review similar work of other authors and formally describe the proposed TNG framework for creating a goal-driven navigation system from separately learnt reactive skills.

%In recent years, deep-learning-based autonomous robot navigation has been looked at from different perspectives, resulting in varied approaches being proposed.
The work from~\cite{8100252}, which includes a mapper and a planner, advances previous approaches. The researchers propose a joint, end-to-end architecture that is differentiable and has the ability to plan with a belief map for the environments where an incomplete set of world observations are provided. A noteworthy feature of ~\cite{8100252} is that their method is able to generalize to new environments. 
Another approach~\cite{SavinovDosovitskiyKoltun2018_SPTM}, proposes a semi-parametric topological memory (SPTM) architecture for navigation, where the environment is mapped on a non-parametric graph, in which every position is a node. In their method  a parametric deep neural network is used to retrieve nodes to reach a specific goal node.
\cite{Bruce2018LearningDN} attempts to tackle the challenge of learning goal-directed policies from data representing a single traversal in an environment spanning over 2 kilometres.
It does so by a topological approach where points in trajectories at a particular distance are represented as nodes and path planning is learnt with random episodic training.

\begin{figure*}[!ht]
	\centering
	\includegraphics[width=0.6\textwidth]{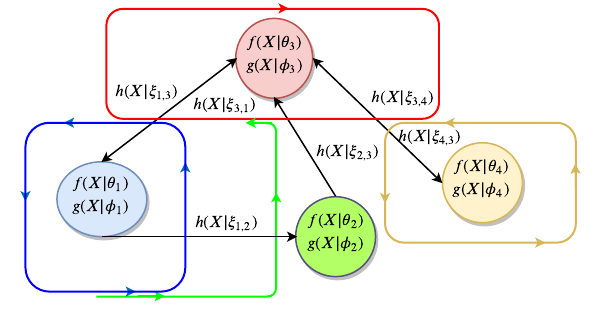}
	\caption{Example of an environment of four intersecting trajectories (directed colored lines) and corresponding TNG. }
	\label{fig:trajectory_graph}
\end{figure*}

The mentioned approaches also have certain limitations.
Whilst end-to-end learning can simultaneously tune the entire architecture's parameter set, this advantage often comes at the cost of requiring large datasets and longer training durations~\cite{LimitsEndToEnd}. Reinforcement learning-based methods such as~\cite{8100252} are dependent on high amounts of resources for training its joint architecture since it needs to tune the parameters by interacting with a well-mapped, photo-realistic environment.
Moreover, the method is also reliant on hard-to-satisfy assumptions, such as the existence of perfect odometry.
While~\cite{Bruce2018LearningDN} is also a reinforcement-learning-based method, to reduce the resource-heavy factor it leverages the method with one-shot reinforcement learning~\cite{bruce2017one}.
While it shows that it can perform planning, the low-level commands that traverse from node to node are controlled by a human operator and are not learnt.
~\cite{SavinovDosovitskiyKoltun2018_SPTM} relies on self-supervised learning but no real-world demonstrations have been shown in the work.

Keeping in mind the above limitations, we propose a structure of perception and action modules, which allows one to plan and perform goal-driven navigation. 

\subsection{Proposed framework}

Let us assume, that the environment is covered by $C \in \mathbb{N}$ trajectories, and let $y = \f$ be trained trajectory imitation controllers (see Section~\ref{sec:trajectory_controllers}), $i=1,2,..,C$, where $X$ is the input image, and $y$ is the control signal for behaviour in the $i$-th trajectory.
We also assume, that the trajectories intersect in such a way that it is possible to reach any given trajectory, starting from any of the aforementioned $C$ trajectories, possibly by switching controllers at intersections.
An example of such environment is depicted in Figure~\ref{fig:trajectory_graph}.
Let us assign to every $i$-th trajectory a binary classifier $\g$ with parameters $\phi_{i}$, which allows to detect that the robot is in this trajectory.
The situations where the robot is at some intersection of the $i$-th and $j$-th trajectories can be recognised by another binary classifier $\h$, where $\xi_{i,j}$ are corresponding parameters.

Utilising the aforementioned components we define TNG as a directed graph $G=(V,E)$, with nodes $V$ corresponding to pairs $(\f, \g)$, and edges $E$ - to $\h$ (see Figure~\ref{fig:trajectory_graph}).

% More clear presentation:
Having TNG and the current sensory input $X_{t}$, we can topologically localise the robot by evaluating $g(X_{t}|\phi_i)$ and $h(X_{t}|\xi_{i,j})$, providing information about the trajectory on which the robot currently is, as well as to which ones it can immediately navigate by starting to execute the corresponding imitation controllers. Similarly, having an externally provided goal image  $X^{*}$ we can identify target trajectory  $i^{\ast}$ by evaluating $i^\ast = \argmax_i g(X^{\ast}|\phi_i)$.

After the shortest path in the TNG between current and target vertices is computed (e.g. by Dijkstra's algorithm \cite{LaValle}), we can start executing the controllers and classifiers, contained in its vertices and edges, switching them at every intersection connecting towards the goal, recognised by corresponding intersection classifiers until the goal trajectory is reached.
Afterwards we continue to execute the goal trajectory controller until the exact location is recognised by the goal-reaching function $r(X_{t}, X^{*}|\psi)$. Here, $\psi$ are corresponding parameters.

%Note, that in this section we omit  the exact functional form of the classifiers, controllers, and description of their training process. We discuss these implementation-related issues in the Section~\ref{sec:implementation}.

\subsection{Notes on Implementation}
\label{sec:implementation}
% HFOV = 70.58 degress

% Mention that TNG is more a framework, that method??

%Further we provide details of our implementation of TNG framework.

%We implemented it on the basis of Neurotechnology's SentiBotics Navigation SDK 3.0, which include an infrastructure for autonomous robot navigation system development and Robot Operating System (ROS) \cite{quigley2009ros} environment.

\subsubsection{Classifiers and goal-reaching function}

Trajectory classifiers $g(X|\phi_{i})$ are constructed using the form mentioned in~\eqref{eq:controller_representation1} with the~\verb|SoftMax| activation of the output layer, as well as $\verb|ReLU|$ activations in the hidden layers.
We use $[512, 256, 32, C]$ neurons in each layer, where $C$ is the number of trajectories in the TNG.
The output of the trajectory classifiers are given by  \begin{equation}
g(X|\phi_{i}) = \mathbb{1}(\argmax g_{\text{MLP}}( g_{\text{MobileNetV2}}(X|\hat{\xi})| \theta) = i),
\end{equation}
where $1 \leq i \leq C$.
The parameters of the classification network, $\theta$, are optimised by minimising cross entropy loss~\cite{murphy2013machine}, and the parameters of the encoder network, $\hat{\xi}$, are pre-trained on the $\text{ImageNet}$~\cite{deng2009imagenet} classification task and kept fixed during training of trajectory imitation controller.
We select $\text{MobileNetV2}$~\cite{MobilenetV2} because it allows us to achieve an acceptable execution speed using low cost embedded neural network inference hardware \footnote{https://software.intel.com/en-us/movidius-ncs}.

Since trajectory intersection classifiers require individual training for each intersection, it adds on a requirement of individually labelling data points for the same.
In order to minimise manual data labelling in our implementation, we perform the learning of intersections with the use of a proprietary object recognition engine derived from~\cite{findobject}, using FAST corners~\cite{FASTRosten} and BRIEF features~\cite{BRIEFCalonder} as descriptors.
This approach allows for the efficient memorisation and recognition of intersections, while being represented by only a few images.
We experimentally found that the same principle proved to be quite efficient to model the goal-reaching function as well.

\subsubsection{Trajectory Imitation Controllers}
We implement TNG with both options of trajectory imitation controllers $f(X|\theta_{i})$, described in Section~\ref{sec:trajectory_controllers}.

\paragraph{Regression Controller.}
In the case of regression controller~\eqref{eq:controller_representation1}, we use MLP architecture, containing \linebreak $[2048,1024,256,32,2]$ neurones and $\verb|ReLU|$ activations in every layer.
The output of the network is clipped at the $[-1.5, 1.5]$ interval, which corresponds to the range of the SentiBotics control pad command.

\paragraph{Object detection-based controller.}
In the case of the object-detection-based controller, we use the architecture described in Section \ref{sec:od_trajectory_controllers}. The controller is trained using bounding boxes of $896\times360$ pixels.
The output of the PID controller is also clipped to the range of the SentiBotics control pad command.

% maybe move to discussion?
\begin{comment}
We have empirically observed that often there is a disproportionate amount of either left-positioned, centered or right-positioned bounding boxes in generated input data.
For instance, when a trajectory mostly contains driving forward actions, the input data is composed of a disproportionate amount of centered rectangles.
This makes the model overfit to one rectangle position, and reduces trajectory following performance.
To tackle this imbalance, we use a bounding box weighting heuristic, which assigns weights as follows: $w_i = n_{items} / (n_{types} * n_i)$, where $i$ (index) corresponds to one of $3$ bounding box position types, $n_{items}$ is the number of items in the training set, $n_{types}$ is the total number of unique bounding box position types in the training set (in our case $3$: left, center, right), and $n_i$ is the $i$-th bounding box position type occurrence count in the training set.
%We weigh each bounding box'es loss function by the its corresponding weight.
\end{comment}

\section{Empirical Evaluation}
\label{sec:experiments}

Our implementation of trajectory imitation controllers and TNG is performed with the use of Neurotechnology's SentiBotics Navigation SDK 3.0~\cite{sentibotics_navigation_sdk}.
We conduct the evaluation in real environments using the $~\text{SentiBotics}$ mobile robot platform prototype, and for simulations, we use Gazebo\footnote{\href{http://gazebosim.org}{http://gazebosim.org}}-based $~\text{SentiBotics}$ simulation modules, from the aforementioned SDK.

We start by describing the evaluation environments and the data collection process, then compare the reactive trajectory imitation controllers described in Section~\ref{sec:trajectory_controller_cnn_mlp} and Section~\ref{sec:od_trajectory_controllers}, and finally evaluate the navigation capability of TNG.

\subsection{Environments and Data}

\begin{figure*}
	\centering
	\subfloat[L1]{\includegraphics[width=1.0\textwidth,height=\textheight,keepaspectratio]{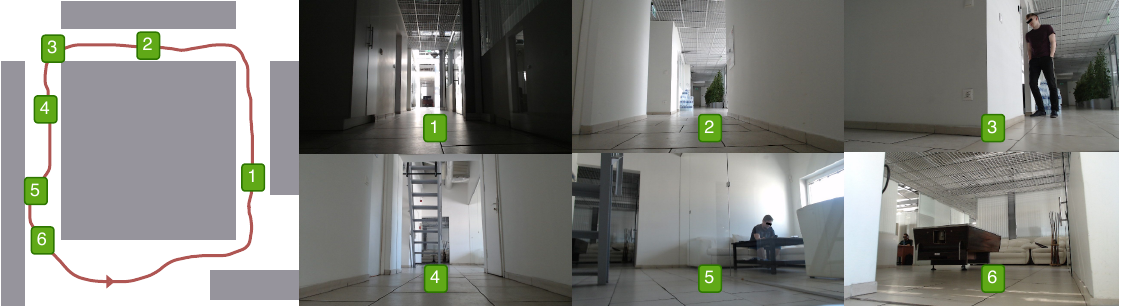}}\hfill
	\subfloat[L2]{\includegraphics[width=1.0\textwidth,height=\textheight,keepaspectratio]{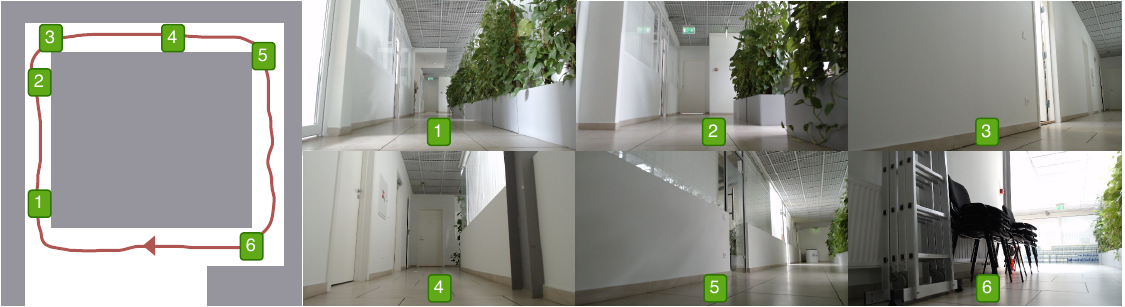}}\hfill
	\subfloat[L3]{\includegraphics[width=1.0\textwidth,height=\textheight,keepaspectratio]{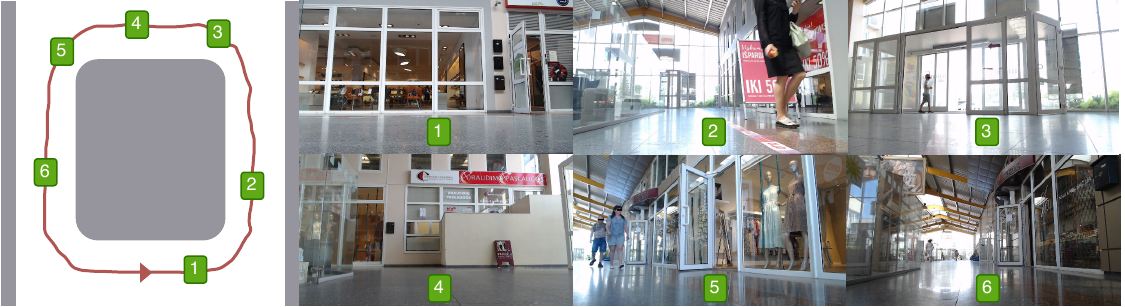}}\hfill
	\subfloat[L4]{\includegraphics[width=1.0\textwidth,height=\textheight,keepaspectratio]{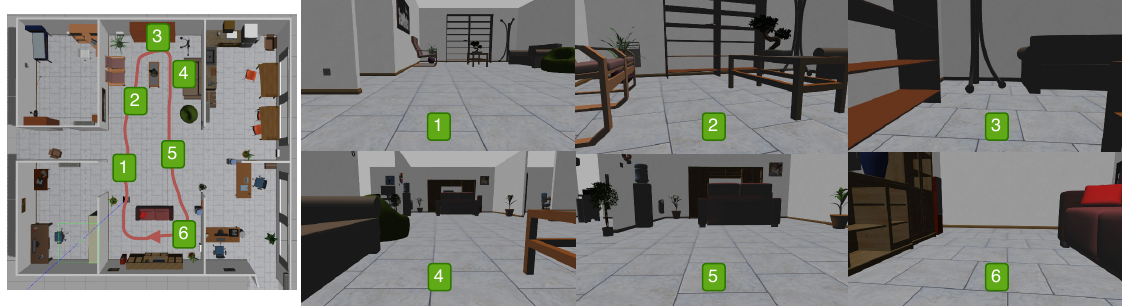}}
	\caption{Environments used to train trajectory imitation controllers on. (a) \& (b) are sitauted inside an office, (c) is situated in shopping mall and (d) is a simulated environment. }
	\label{fig:traj_imitation_envs}
\end{figure*}

To evaluate trajectory imitation controllers, we utilise three real-world environments and one simulated environment.
These are closed-loop trajectories where two are situated inside an office (referred to as L1 and L2), one is situated in a shopping centre (L3) and the simulated environment (L4) consists of a large room with objects such as couches and tables.
The trajectories are depicted in Figure ~\ref{fig:traj_imitation_envs}.
In turn, for the evaluation of the navigation capabilities of TNG, we utilised an office environment with multiple connecting corridors, and the map can be seen in Figure~\ref{fig:trajectory_graph_experiments}.
The real-world environments are dynamic in terms of lighting since they are open to natural light and data is collected during the day. In addition, there are pedestrians walking around during the data collection and experimentation, while the simulated environment is completely static.

\begin{figure}[t]
    \centering
	\includegraphics[width=0.50\textwidth]{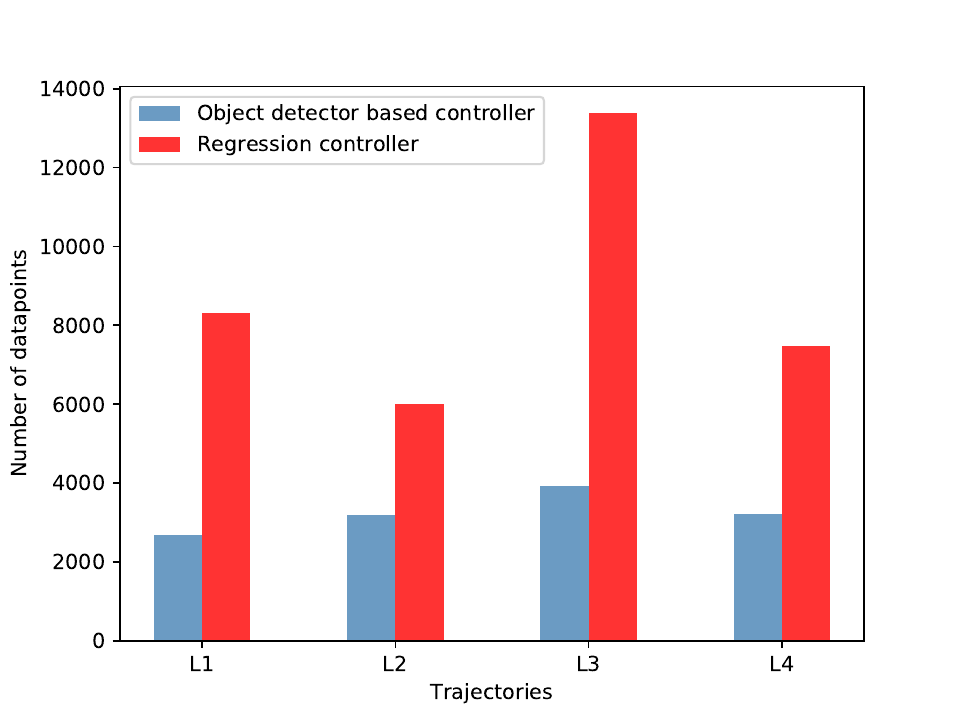}
	\caption{Number of data points used to achieve reported scores}
	\label{fig:datapoints}
\end{figure}

To collect the data for each of the trajectories to be learnt, we drive the robot through each of the trajectory three times and record the images seen from the robot's front facing RGB camera, as well as the corresponding motor commands.

\subsection{Experiments with Trajectory Imitation Controllers}
\label{sec:experiments_trajectory_controllers_comparison}

We train both the controllers in the ways mentioned earlier, where the base data remains the same, but the regression controller contains aggregated data from a few iterations of DAgger.
The number of iterations is dependent on the complexity of the trajectory to learn.

Once both the controllers have been trained for a trajectory, we run the controller on the closed trajectory with a target of completing 10 laps.
We observe the execution actively and intervene only when we notice the controller had made a mistake.
Each time there is an intervention, we record the time required to correct the robot and bring it back to the trajectory.
The experiments for evaluation are performed twice, first immediately after data collection and subsequently after a period of 4 weeks after data collection.
This step is done to check if the performance holds up after there have been changes in the environments, hence we also compare the differences between the two evaluations.
We do not repeat the experiments in the simulated environment since it is static and not bound to change with time.

To compare the performance of trajectory imitation controllers we use percentage autonomy~\cite{BojarskiTDFFGJM16} as a controller quality measure:

\begin{equation}
PA = 100\cdot (1-\frac{\tau_{h}}{\tau}),
\label{eq:percentage_autonomu}
\end{equation}
where $\tau_{h}$ is duration of time when the robot is controlled by a human (i.e. at times where correction is needed), and $\tau$ denotes the total duration of execution.

%\small
\begin{table}
 
    \begin{center}
    	\begin{tabular}{|c|c|c|c|c|}
    		\hline
    		Trajectory & Controller & PA (\%) & PA (\%)  &  Difference \\
    		name & type & before  & after  & (\%) \\
    		\hline
    		L1 & Reg. & 99.9 & 97.0 & 2.9 \\
    		L2 & Reg. & 99.8 & 98.7 & 1.1 \\
    		L3 & Reg. & 99.9 & 99.3 & 0.6 \\
    		\hline
    		L1 & Obj. det. & 98.0 & 97.1 & 0.9 \\
    		L2 & Obj. det. & 99.7 & 99.5 & 0.2 \\
    		L3 & Obj. det. & 99.6 & 99.3 & 0.3 \\
    		\hline
    		L4 & Reg. & 99.9 & - & - \\
    		L4 & Obj. det. & 100.0 & - & - \\
    		\hline
%    		\hline
    	\end{tabular}
    	\caption{Percentage autonomy scores calculated before and after an interval of 4 weeks.}
    	\label{table:controller_comparison_results}
    \end{center}
\end{table}
%\normalsize 

{\bf Discussion.} Table \ref{table:controller_comparison_results} displays the percentage autonomy score obtained for each trajectory before and after the interval, along with the difference in scores.
Both controllers perform nearly perfectly in the simulated environment, possibly due to its static nature, while in the real-world environments, the performance is comparatively lower.
Over the two evaluations over intervals, the regression controller demonstrates higher degradation in performance than the object-detection-based controller.

We speculate that the reason for the regression controller to have a higher score in the first evaluation could be due to the several DAgger iterations performed on every trajectory's controller to improve until almost perfection, causing to over-fit to the appearance of the environment at the time of the last DAgger iteration.
In contrast, in the second evaluation, the appearance goes through changes due to environmental conditions, as tests were carried out at a different time of the day.

% DEFINE "INITIAL DATASET" in some previous section, referred to as base data before, agree on something
The object detection controller's setting poses trajectory imitation as a problem of detecting the direction of the path rather than regressing direct commands given an image, making it simple enough to produce commands heuristically.
The performance of this setting indicates to be better utilizing the initial dataset and does not need aggregated data to reach high autonomy, compared to the regression controller for which it takes several DAgger iterations to reach similar autonomy. %, as shown in Figure~\ref{fig:traj_imitation_envs}.
With lower differences in the percentage autonomy between evaluations before and after an interval, object detection based controller shows better tolerance to environmental changes, while creating fewer possibilities for cascading errors to occur.

We observed that the regression controllers' mistakes were often unpredictable where as object detection based controllers made mistakes at specific locations in the trajectories.
Hence, the performance can be further enhanced by exploring the application of DAgger on the object detection-based controller.
We also noticed the object detection based controllers with better loss over the test sets show better real-world performances, which is not the case for the regression controller~\cite{Codevilla2018OnOE}, this can be further investigated in future work.

\subsection{Experiments with TNG}
\label{sec:navigation_system_experiments}

\begin{figure}
\centering
\includegraphics[width=0.60\textwidth]{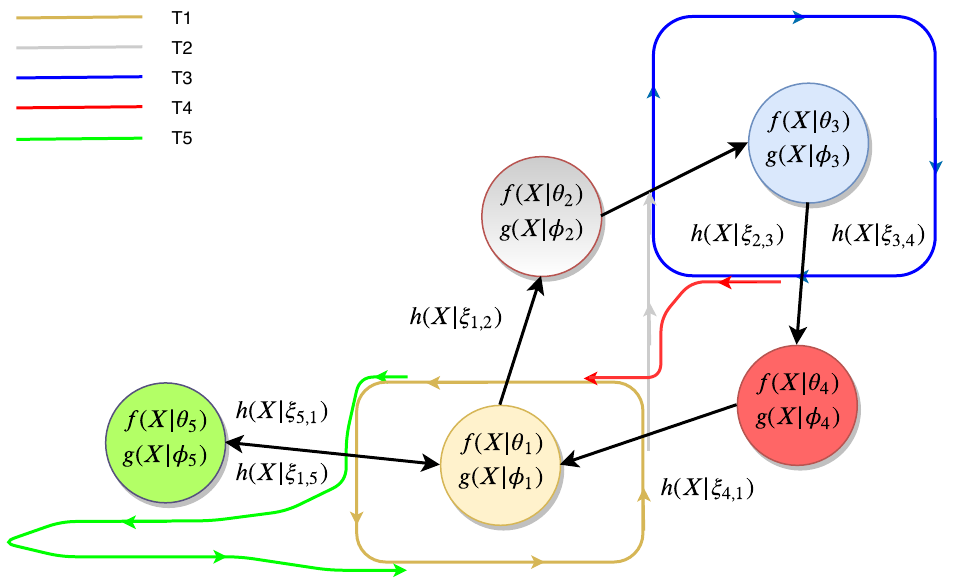}
\caption{Schematic of experiment trajectories T1-T5 depicted in different colors, and corresponding TNG model.}
\label{fig:trajectory_graph_experiments}
\end{figure}

\begin{figure}
	\centering	
	\includegraphics[scale=0.42]{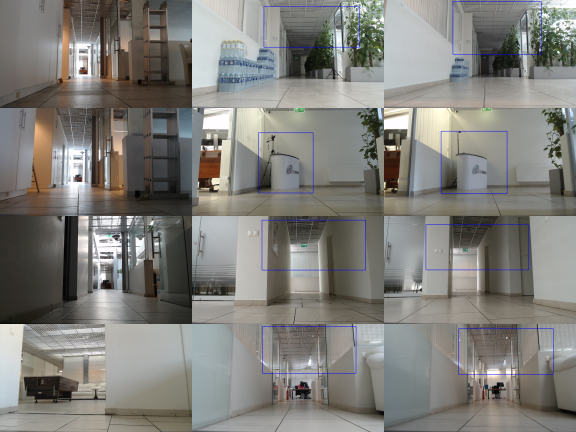}
	\caption{Results of autonomous navigation from trajectory T1 to the goals from the remaining trajectories (T2,T3,T4,T5).}
	\label{fig:tng_experiments1}	
\end{figure}

\begin{figure}
\centering
\includegraphics[scale=0.42]{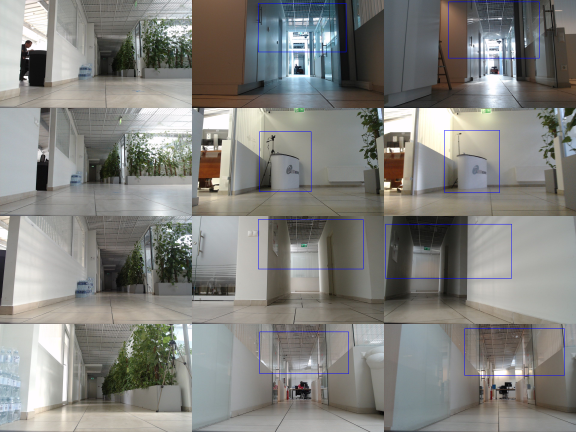}
\caption{Results of autonomous navigation from trajectory T2 to the goals from the remaining trajectories (T1,T3,T4,T5).}
\label{fig:tng_experiments2}
\end{figure}

\begin{figure}
\centering
\includegraphics[scale=0.42]{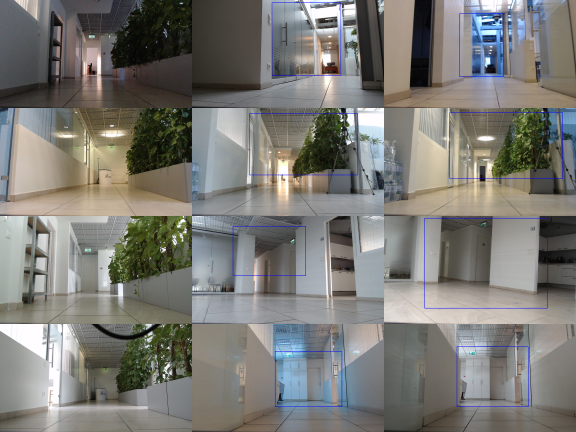}
\caption{Results of autonomous navigation from trajectory T3 to the goals from the remaining trajectories (T1,T2,T4,T5). }
\label{fig:tng_experiments3}	
\end{figure}

\begin{figure}
\centering
\includegraphics[scale=0.42]{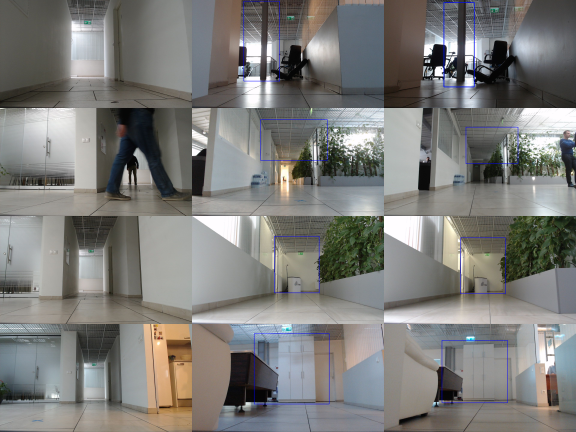}
\caption{Results of autonomous navigation from trajectory T4 to the goals from the remaining trajectories (T1,T2,T3,T5). }
\label{fig:tng_experiments4}
\end{figure}

\begin{figure}
\centering
\includegraphics[scale=0.42]{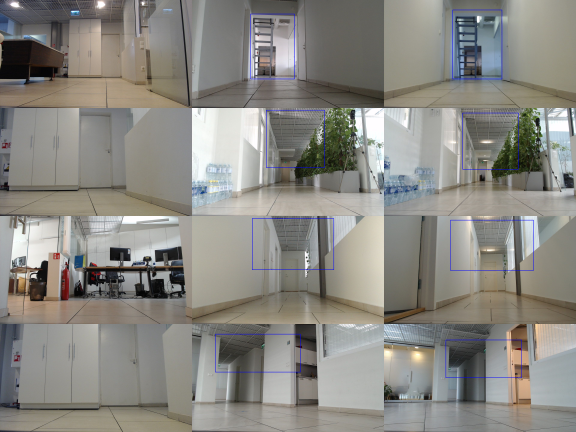}
\caption{Results of autonomous navigation from trajectory T5 to the goals from the remaining trajectories (T1,T2,T3,T4).}
\label{fig:tng_experiments5}	
\end{figure}

We cover a subset of an office environment with five trajectories (denoted as T1-T5), and construct the corresponding TNG, as depicted in Figure~\ref{fig:trajectory_graph_experiments}.
For each different trajectory pair, we randomly selected initial and goal positions, and we conduct autonomous navigation episodes between them, recording average percentage autonomy, travelled distance in metres (Table~\ref{table:tng_stats}), as well as corresponding initial, goal, and final images, which provide information on the accuracy of each navigation experiment. These images are depicted in Figures~\ref{fig:tng_experiments1},~\ref{fig:tng_experiments2},~\ref{fig:tng_experiments3},~\ref{fig:tng_experiments4} and~\ref{fig:tng_experiments5} in the first, second and third columns respectively. Blue rectangles correspondingly denote visually specified and recognised navigation goal. We conduct  all the experiments in a natural office environment during working hours.

The obtained empirical results indicate that the TNG framework indeed allows one to utilise pre-trained reactive trajectory imitation controllers to achieve goal-directed navigation in real-world environments.

% total traveled distance = 1119,68m
\begin{table}
	\begin{center}
		\begin{tabular}{|c|c|c|c|c|c|}						
      \hline
      & \multicolumn{4}{c}{Percentage autonomy} & \\ 			\hline
			& T1 & T2 & T3 & T4 & T5  \\
			\hline 
			T1 & - & 99.1 & 96.4 & 93.4 & 92.9 \\
			T2 & 96.0 & - & 96.3 & 98.2 & 97.9 \\
			T3 & 94.9 & 92.1 & - & 94.6 & 98.3\\
			T4 & 99.1 & 97.0 & 97.2 & - & 97.3  \\
			T5 & 95.8 & 95.9 & 98.7 & 94.6 & - \\
			\hline
    \end{tabular}
    \begin{tabular}{|c|c|c|c|c|c|}						
      \hline
      &  \multicolumn{4}{c}{Distance (m.)} &\\			\hline
			& T1 & T2 & T3 & T4 & T5  \\
			\hline 
			T1 &  - & 18.5 & 65.6 & 75.2 & 8.5\\
			T2 &  85.3 & - & 47.7 & 62.2 & 82.2\\
			T3 &  80.1 & 71.6 & - & 59.9 & 73.8\\
			T4 &  14.8 & 38.2 & 81.1 & - & 39.2 \\
			T5 &  27.2 & 31.5 & 61.0 & 96.1 & -\\
			\hline
    \end{tabular}
    
		\caption{Statistics of TNG navigation experiments. Rows and columns correspond to the source and destination trajectories respectively. On the left we report averaged percentage autonomy values (mean $\approx 96.3$), and on the right we provide traveled distance in metres (total distance $\approx 1.1$ kilometres).}
		\label{table:tng_stats}
	\end{center}
\end{table}

% \subsection{Discussion}

{\bf Discussion.} In contrast to metric models (e.g.~\cite{ORBSLAM,8100252,LSDSLAM}), where localisation is performed by geometrically estimating a robot's pose on a map, or to other topological  (e.g.~\cite{SavinovDosovitskiyKoltun2018_SPTM, Fraundorfer2007TopologicalML,Bruce2018LearningDN}) or hybrid (e.g.~\cite{ThrunLearningTopological,RatSLAM}) approaches, the trajectory classifiers and trajectory intersection classifiers of TNG provide only minimal localisation information, required to keep or switch reactive behaviour in a way that eventually leads to reaching the given goal.
The modularity of the proposed TNG framework offers two important advantages. First one is that it allows one to augment TNG incrementally by adding to the graph the required imitation controllers and classifiers, which correspond to the included additional navigation area. And the second advantage is that the perception (trajectory and trajectory intersection recognition) and action (trajectory imitation) modules of TNG can be selected from the results of visual classification/recognition and  reactive imitation learning respectively, which potentially makes TNG attractive both for the robotics researchers and practioners.

Although the TNG framework is composed of "black box" modules, its operation is transparent and easily interpretable at the system level, which can also be regarded as a practical advantage, compared to entirely "black box" models (e.g.~\cite{8100252}). 

On the other side, the proposed TNG approach assumes that the navigation environment should consist of separable, intersecting trajectories. Therefore, it may not cope well with arbitrary environment coverings (e.g., two almost parallel nearby trajectories would be hard to discriminate). 
Moreover, TNG is geometrically sub-optimal, and requires significant human effort - both when collecting training data and when designing the navigation graph.
However, there are many realistic scenarios (e.g. navigation in buildings and roads), where the  aforementioned intersecting trajectory environment assumption is naturally satisfied.

\section{Conclusion}
\label{sec:conclusion}

% Practical usefulness!

In this article, we studied how to utilise reactive trajectory imitation controllers for topological, goal-directed autonomous robot navigation. As a solution, we suggested a novel TNG framework (see Figure~\ref{fig:trajectory_graph}). We also proposed a novel application of neural object detection architectures for visuomotor trajectory imitation. We performed an empirical evaluation of the suggested algorithms both in a simulator and in reality. The experiments conducted reveal that neural object detection architectures can be efficiently applied for visuomotor trajectory imitation, and the proposed TNG framework allows to compose reactive trajectory imitation modules into a goal-directed navigation system capable of achieving visually specified goal states. 

Our future work will include research on more efficient and reliable imitation learning approaches able to handle more dynamic environments, optimising TNG by adding a fully automatic environment graph construction capability and extending it with a hierarchical component.

\section*{Funding}
\noindent This research has been funded by Neurotechnology.

\section*{Conflict of interest}
% https://www.ncbi.nlm.nih.gov/pmc/articles/PMC4219433/ ???
The authors declare that the research was conducted in the absence of any commercial or financial relationships that could be construed as a potential conflict of interest.

\section*{Acknowledgements}
We are grateful to Neurotechnology for providing resources
and support for this research.

\bibliographystyle{unsrt}

\bibliography{bibliography}

\end{document}